\documentclass[sigconf]{acmart}

\usepackage{multirow}
\usepackage{makecell}
\usepackage{colortbl}
\usepackage[utf8]{inputenc} 
\usepackage[T1]{fontenc}    
\usepackage{hyperref}       
\usepackage{url}            
\usepackage{booktabs}       
\usepackage{amsfonts}       
\usepackage{nicefrac}       
\usepackage{microtype}      
\usepackage{xcolor}         
\usepackage{wrapfig,lipsum,booktabs}
\usepackage{soul}
\usepackage{multirow}
\usepackage{algorithm}
\usepackage{algpseudocode}%

\newlength\myindent
\usepackage{amsmath}
\usepackage{mathtools}
\usepackage{amsthm}
\usepackage{subcaption}

\author{Arezoo Rajabi}
\authornote{Equal contribution}
\affiliation{%
  \institution{University of Washington}
  \streetaddress{University of Washington}
  \city{Seattle}
  \state{}
  \country{USA}}
\email{rajabia@uw.edu}

\author{Reeya Pimple}
\authornotemark[1]
\affiliation{%
  \institution{University of Washington}
  \streetaddress{University of Washington}
  \city{Seattle}
  \state{}
  \country{USA}}
\email{reeyabp@uw.edu}

\author{Aiswarya Janardhanan}
\affiliation{%
  \institution{University of Washington}
  \streetaddress{University of Washington}
  \city{Seattle}
  \state{}
  \country{USA}}
\email{ajanard5@uw.edu}

\author{Surudhi Asokraj}
\affiliation{%
  \institution{University of Washington}
  \streetaddress{University of Washington}
  \city{Seattle}
  \state{}
  \country{USA}}
\email{surudh22@uw.edu}

\author{Bhaskar Ramasubramanian}
\affiliation{%
  \institution{Western Washington University}
  \streetaddress{Western Washington University}
  \city{Bellingham}
  \state{}
  \country{USA}}
\email{ramasub@wwu.edu}

\author{Radha Poovendran}
\affiliation{%
  \institution{University of Washington}
  \streetaddress{University of Washington}
  \city{Seattle}
  \state{}
  \country{USA}}
\email{rp3@uw.edu}

\renewcommand{\shortauthors}{Arezoo Rajabi et al.}

\title{Double-Dip: Thwarting Label-Only Membership Inference Attacks with Transfer Learning and Randomization}
\begin{document}
\renewcommand{\shorttitle}{Double-Dip}
\begin{abstract}
Transfer learning (TL) has been demonstrated to improve DNN model performance when faced with a scarcity of training samples. 
However, the suitability of TL as a solution to reduce vulnerability of overfitted DNNs to privacy attacks 
is  unexplored. 
A class of privacy attacks called membership inference attacks (MIAs) aim to determine whether a given sample belongs to the training dataset (\emph{member}) or not (\emph{nonmember}). 
We introduce \textbf{Double-Dip}, a systematic empirical study investigating the use of TL (Stage-1) combined with randomization (Stage-2) to thwart MIAs on overfitted DNNs without degrading classification accuracy. 
Our study examines the roles of shared feature space and parameter values between \emph{source} and \emph{target} models, number of frozen layers, and complexity of \emph{pretrained} models. 
We evaluate Double-Dip on three (Target, Source) dataset pairs: (i) (CIFAR-10, ImageNet), (ii) (GTSRB, ImageNet), (iii) (CelebA, VGGFace2). 
We consider four publicly available pretrained DNNs: (a) VGG-19, (b) ResNet-18, (c) Swin-T, and (d) FaceNet. 
Our experiments demonstrate that Stage-1 reduces adversary success while also significantly increasing classification accuracy of nonmembers against an adversary with either white-box or black-box DNN model access, attempting to carry out SOTA label-only MIAs. 
After Stage-2, success of an adversary carrying out a label-only MIA is further reduced to near $50\%$, bringing it closer to a random guess and showing the effectiveness of Double-Dip. 
Stage-2 of Double-Dip also achieves lower ASR and higher classification accuracy than regularization and differential privacy-based methods. 
\end{abstract}
\maketitle
\section{Introduction}\label{sec:Introduction}

Deep neural networks (DNNs) leverage 
cost-effective storage and computing to achieve high classification accuracy on multiple data-intensive applications (e.g., face-recognition~\cite{taigman2014deepface}, disease diagnosis~\cite{esteva2019guide}). 
The ability of DNNs to classify previously unseen inputs with high accuracy relies critically on being trained on large datasets, and requires significant training-time computational resources~\cite{bender2021dangers, birhane2021large}. 
Use of 
public pretrained models~\cite{he2016deep, simonyan2015very} and online 
platforms~\cite{AWS, BigML, Caffe} have been proposed as  solutions to reduce costs and training time incurred by users having to train their own models.

\begin{figure*}[!h]
\centering 
    \includegraphics[width=0.9\textwidth]{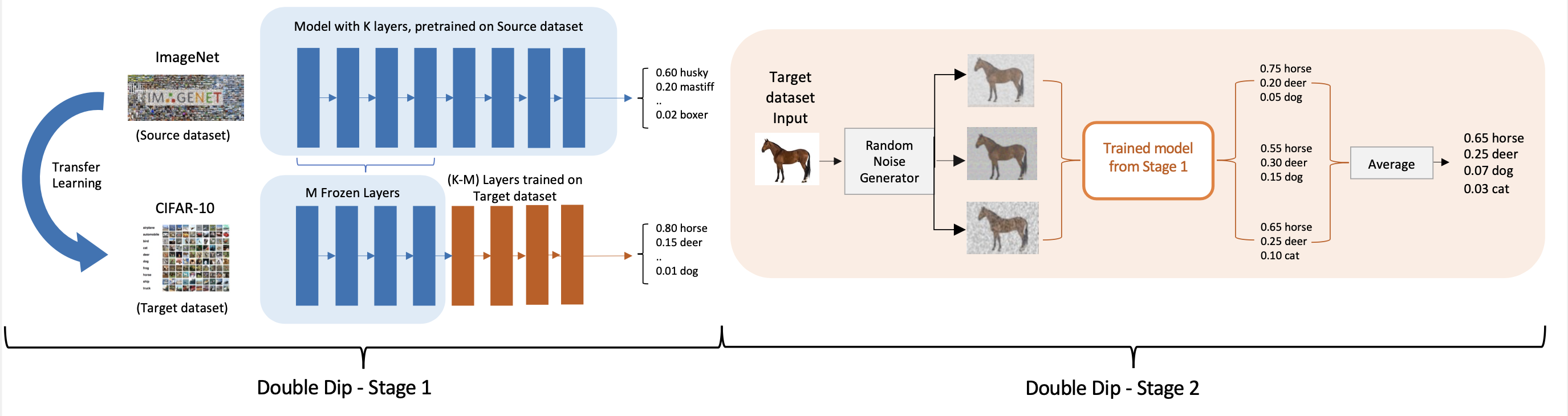}
    \caption{\textbf{Double-Dip Mechanism}. Stage-1 uses transfer learning to embed features of a lower dimensional overfitted DNN into a \emph{target model} that overcomes overfitting. The target model is learned by `freezing' weights in $M$ layers of a publicly available pretrained model, and using samples from the target dataset to learn weights of the remaining $K-M$ layers of the pretrained model. Overcoming overfitting will enable resilience to membership inference attacks (MIAs) by reducing success rate of an adversary even when the size of the training dataset for the target model is limited. 
    Stage-2 employs randomization to generate multiple noisy variants of a given input sample $x$. 
    Each noisy variant is provided to the trained target model from Stage-1 to obtain the possible output class labels as probabilities. 
    An averaging mechanism is then used to `smooth' these output class labels to obtain the final output class label. 
    The key insight underpinning Stage-2 is that randomization will affect estimates of the distance of a data point to a decision boundary. As a result, the final output label $y$ will not reveal information about whether the input $x$ was used to train the target model (member) or not (nonmember). 
    }
    \label{fig:DoubleDipSchematic}
\end{figure*}

Although pretrained models and online platforms help alleviate costs of training, two challenges often arise. 
First, the dataset used to train a pretrained DNN might be different from the dataset of interest that belongs to a potential user of the model. 
The user may then have to \emph{retrain} (weights in some layers of) the DNN~\cite{zhuang2020comprehensive}. 
Second, the user may have access to only a limited number of data samples, or might not be willing to provide large amounts of data to (re)train the model. 
In the absence of an adequate number of training samples, the DNN model can suffer from \emph{overfitting}~\cite{hastie2009elements}. 
Overfitted DNNs have been shown to `memorize' patterns in the data and classify samples belonging to the training dataset with high accuracy, while performing poorly on other samples~\cite{shokri2017membership}. 

Overfitted DNNs have been shown to be vulnerable to privacy attacks~\cite{liu2021machine}. 
One instance of a privacy attack due to overfitting is realized as a \emph{membership inference attack (MIA)}~\cite{shokri2017membership}. 
%
MIAs aim to determine if a given sample of interest belongs to the training dataset (\emph{member}) of a DNN model or not (\emph{nonmember})~\cite{jia2019memguard, rajabi2022ldl, yeom2018privacy}. 
MIAs 
can result in disclosure of sensitive information (e.g., social-security numbers~\cite{carlini2019secret}), resulting in privacy threats. 
An adversary carrying out a MIA uses adversarial learning \cite{goodfellow2014FGS, BIM} to estimate the smallest magnitude of noise, denoted $\delta$, that needs to be added to a given sample so that the DNN model misclassifies this sample. 
The value of $\delta$ enables an adversary to distinguish between members and nonmembers based on a heuristic that member samples are typically located relatively farther from a decision boundary and are robust to noise perturbations compared to nonmembers~\cite{choquette2021label, rajabi2022ldl}. 
This type of MIA only uses the predicted output label of the sample from the DNN without requiring 
confidence scores associated to labels, and has been called a \emph{label-based} or \emph{label-only} MIA 
\cite{choquette2021label, li2021membership, rajabi2022ldl}. 

Techniques including differential privacy~\cite{abadi2016deep}, regularization~\cite{nasr2018machine}, and distillation~\cite{tang2022mitigating} have been used as defenses against MIAs. 
However, these methods have also been shown to lower classification accuracy for overfitted DNNs~\cite{rajabi2022ldl}, which can affect usability of the model. 
Further, their effectiveness on label-only MIAs is less understood. 
Finding solutions to mitigate impacts of label-only MIAs while improving classification accuracy for overfitted DNNs remains an open problem. 

\noindent{\bf Our Contribution:} In this paper, we propose \textbf{Double-Dip}, a systematic empirical study of using transfer learning (TL) to overcome overfitting in the limited data setting, thus resulting in thwarting of label-only MIAs. 
While the usefulness of TL in the general limited data setting is well-known, we show in this paper that TL will indeed be helpful even in the case of overfitted DNNs. 

In Double-Dip Stage-1, we demonstrate that TL \cite{zhuang2020comprehensive} will help embed an otherwise low-dimensional overfitted model into a high-dimensional \emph{target model} that will be less overfitted. 
Our key insight is that overcoming overfitting will enable resilience to MIAs by reducing success of an adversary even when size of the training dataset for the target model is limited. Using transfer learning in the limited data setting will also enhance model usability, characterized by an increase in classification accuracy of nonmember samples. However, transfer learning alone may not be adequate to reduce success of an adversary carrying out label-only MIAs. 

In Stage-2, we employ randomization to construct a region 
of constant output label centered at a given input sample such that the DNN model returns the same output label for all data points inside this region  \cite{cohen2019certified, rajabi2022ldl, ye2022one}.  
The intuition underpinning Stage-2 is that such randomization will affect estimates of the distance of a data point to a decision boundary. 
As a consequence, magnitudes of noise required to misclassify member and nonmember samples will be comparable, thereby preventing a querying adversary from distinguishing between members and nonmembers. 
Stage-2 will help further reduce success rate of an adversary carrying out a label-only MIA, which is the most powerful known MIA to date~\cite{choquette2021label}, without reducing classification accuracy (relative to Stage-1).  

Together, the two stages will help reduce success of an adversary carrying out a label-only MIA while also yielding a target model with high accuracy. 
Fig. \ref{fig:DoubleDipSchematic} illustrates the mechanism of Double-Dip. 

\noindent
\textbf{Evaluation:} 
To evaluate Double-Dip, we consider an adversary carrying out a SOTA label-only MIA~\cite{choquette2021label}. 
We perform extensive experiments to evaluate classification accuracy and success rate of an adversary by examining roles of (i) shared features between datasets used to train pretrained and target models, (ii) the complexity of the pretrained model, (iii) freezing different numbers of layers of the pretrained model for different sizes of the target dataset, and (iv) different defense mechanisms against MIAs. 

We examine multiple datasets to learn target models using a pretrained model trained on a different source dataset. 
We consider the following (Target, Source) dataset pairs: (i) (CIFAR-10, ImageNet), (ii) (GTSRB, ImageNet), (iii) (CelebA, VGGFace2). 
We consider different publicly available pretrained DNNs, including VGG-19~\cite{simonyan2015very}, ResNet-18~\cite{he2016deep}, Swin-T~\cite{liu2021swin}, and FaceNet~\cite{schroff2015facenet}. 
Our choice of pretrained DNN models encompasses distinct architecture types, including convolutional neural networks (CNN) (VGG-19, ResNet-18), transformers (Swin-T) for image recognition, and a CNN model designed for face recognition (FaceNet). 
We propose randomization based on noise perturbation around an input sample to construct a region of constant output label. We also compare the impact of mechanisms used to reduce success of an adversary carrying out a MIA by investigating regularization \cite{nasr2018machine}, distillation \cite{tang2022mitigating}, and differential privacy \cite{abadi2016deep} with Double-Dip.

\noindent 
\textbf{Key Results:} 
Our results reveal that for overfitted DNNs, using TL in Stage-1 consistently reduces the success rate of an adversary carrying out a SOTA label-only MIA~\cite{choquette2021label}. 
For example, when using $500$ samples from the GTSRB dataset to learn a target model using a ResNet-18 pretrained model, Stage-1 of Double-Dip achieves an ASR of \textbf{54.1\%}, compared to 77.2\% without TL. 
Simultaneously, Stage-1 significantly improves classification accuracy for nonmember samples by up to \textbf{3.4x} higher than SOTA regularization-based~\cite{nasr2018machine} and \textbf{2.6x} higher than distillation-based~\cite{tang2022mitigating} methods. 
After Stage-2, we observe that the success rate of an adversary carrying out a label-only MIA is reduced to $\mathbf{50.0\%}$ while maintaining classification accuracy of nonmembers. 
Furthermore, our proposed randomization-based strategy for Stage-2 is accomplished by a lightweight post processing module that does not require retraining of target DNN models. 
These results underscore the effectiveness of Double-Dip in thwarting an adversary carrying out label-only MIAs. 

The rest of the paper is organized as: 
Sec. \ref{sec:preliminaries} introduces preliminaries. 
Sec. \ref{sec:ThreatModel} presents the threat model. 
The working of Double-Dip is detailed in Sec. \ref{sec:methodologyDoubleDip}. 
Sec. \ref{sec:Evaluation} gives our evaluation setup, and we present results in Sec. \ref{sec:evaluation}. 
We discuss salient features of Double-Dip in Sec. \ref{sec:Discussion} and describe related work in Sec. \ref{sec:RelatedWork}. 
Sec. \ref{sec:conclussion} concludes the paper. 
\section{Preliminaries} \label{sec:preliminaries}
This section describes needed background on overfitted DNNs and TL, and specifies metrics used to evaluate our approach. \\

\noindent 
\textbf{Overfitted DNNs}: 
A DNN classifier is said to be \emph{overfitted} if the model classifies members (samples in training set) correctly with high confidence while having lower output label accuracy of classification when identifying nonmembers~\cite{shokri2017membership}. 
The high confidence of an overfitted model on member samples is because the model `memorizes' patterns in the training dataset and is unable to generalize to `unseen' data samples (nonmembers). 
This phenomenon is exacerbated in DNNs with a large number of tunable parameters, where the model can tend to learn too many details associated with training data along with any noise that might be present in the training data. 
In~\cite{jia2019memguard, yeom2018privacy}, it was shown that overfitting of DNNs during training makes the model highly vulnerable to MIAs.\\

\noindent 
\textbf{Transfer Learning}: 
%
Although DNNs can be trained to achieve high accuracy on a variety of tasks, the training procedure can be computationally expensive. 
The dependence of DNN model training on possession of large training datasets can be partially alleviated through \emph{transfer learning} \cite{zhuang2020comprehensive}. 
The starting point for transfer learning is a (publicly available) \textbf{pretrained model} that has been trained on a \textbf{source dataset}. 
Selected pretrained models are trained on a dataset different from, but often related to the \textbf{target dataset}, which is the given dataset of interest belonging to a user. 
The target dataset is used to learn parameters of the \textbf{target model}. 
Transfer learning has been shown to reduce computational costs of learning the target model, utilizing the pretrained model and \emph{freezing} a suitable subset of layers of the pretrained model~\cite{zhuang2020comprehensive}. 
The target dataset is then used to only train the remaining layers of the DNN~\cite{zhuang2020comprehensive}. \\

\noindent 
\textbf{Metrics}: 
One way to quantify effectiveness of a MIA is to count the number of data samples, denoted by $x$, that are identified correctly as members or nonmembers. 
Samples that belong to the training dataset of the model (member set, $S_m$) are assumed to have membership label $s=1$. 
Any other sample (nonmember set $S'_m$) has membership label $s=0$.   
\emph{True positive rates} (TPR) and \emph{true negative rates} (TNR) indicate correct classification rates for members and nonmembers, and are defined as 
$TPR = Pr\{s=1; x \in S_m\}$, $TNR = Pr\{s=0; x \in S'_m\}$. 
The \textbf{adversary success rate (ASR)}, defined as $ASR = 0.5 (TPR+TNR)$~\cite{shokri2017membership}, measures the average accuracy of a MIA in distinguishing members from nonmembers. 
An \textbf{ASR value closer to $50\%$}- i.e., a random guess- is an indicator of the adversary \emph{not being able to effectively distinguish between member and nonmembers}. 
\section{Threat Model}\label{sec:ThreatModel}
In this section, we introduce the threat model, and describe assumptions, goals, and capability of the adversary.

\noindent
\textbf{Adversary Assumption and Goals}: 
The adversary is assumed to have adequate data samples and computational resources, and uses a SOTA \emph{label-only MIA}~\cite{choquette2021label} to determine if a given input sample is contained in the set that is used to train the model (\emph{member}). 
Label-only MIAs distinguish members from nonmembers based on the minimum magnitude of noise that needs to be added to a sample in order for the DNN to misclassify the sample~\cite{choquette2021label, rajabi2022ldl}. 
The magnitude of noise enables an adversary to distinguish between members and nonmembers based on a heuristic that members are relatively farther away from a decision boundary and are robust to small noise perturbations compared to a nonmember \cite{choquette2021label, jia2019memguard, rajabi2022ldl}. 

\noindent
\textbf{Adversary Actions}: 
Effectiveness of an adversary carrying out an MIA depends on the information it has about the target model~\cite{carlini2019secret}). 
We consider two levels of access to the target DNN model for the adversary: (i) white-box access, where the adversary has access to both, model hyperparameters and output labels, and (ii) black-box access, where the adversary has access only to model outputs. 

\noindent
\emph{White-Box Access}: 
An adversary with white-box access uses an adversarial learning method, e.g., \emph{basic iterative method (BIM)}~\cite{BIM}, to estimate a threshold $\delta$ on noise needed to be added to a sample for it to be misclassified by the DNN. 
The BIM is a SOTA computationally inexpensive algorithm involving repeated application of a fast-gradient sign (FGS) method~\cite{goodfellow2014FGS}, where FGS generates adversarial noise in direction of the gradient of a loss function. 

\noindent 
\emph{Black-Box Access}: 
An adversary with black-box access uses a recently introduced query-based SOTA adversarial learning method (e.g., \emph{HopSkipJump}~\cite{chen2020hopskipjumpattack}) to estimate $\delta$. 
In some cases, an adversary with black-box access may not have adequate number of samples to estimate $\delta$. However, it may have access to a sufficiently large dataset of other samples along with their labels, which can be used to train \emph{shadow or substitute models}~\cite{choquette2021label, rajabi2022ldl} that are similar to the target model. 
The shadow models can then be used to estimate $\delta$. 
\section{Double-Dip: A Two-Stage Approach}\label{sec:methodologyDoubleDip} 
In this section, we describe the two-stage procedure used by \textbf{Double-Dip}. 
The performance of Double-Dip will be assessed in terms of the adversary success rate (ASR- \emph{closer to 50.0\% is better}) and classification accuracy of nonmembers (ACC- \emph{higher is better}). 
Stage-1 uses transfer learning (TL)~\cite{zhuang2020comprehensive} to embed a lower dimensional DNN model into a high-dimensional target model to overcome overfitting. 
Stage-2 employs randomization based on noise perturbation of a given data sample to construct a high-dimensional region of constant output label such that the DNN returns the same label for every data point within this sphere \cite{cohen2019certified, rajabi2022ldl, ye2022one}. 

\subsection{Stage-1 of Double-Dip}
When a user possesses only a limited number of samples to train a DNN, the resulting model becomes \emph{overfitted}, lowering classification accuracy for nonmembers while having high accuracy for members \cite{hastie2009elements}. 
Overfitted DNNs are known to be vulnerable to MIAs \cite{shokri2017membership}. 
Stage-1 of Double-Dip aims to overcome overfitting by using transfer learning, resulting in a target model with increased classification accuracy for nonmembers. 
Our insight is that TL helps embed an otherwise low-dimensional overfitted model into a high-dimensional model that will no longer be overfitted. 
The success of Stage-1, however, will depend on an interplay among several design choices, including the type of pretrained model, source and target datasets, and number of frozen layers of the pretrained model. 
We pose the following questions that form the basis of evaluating performance of Stage-1 of Double-Dip 
for overfitted DNNs.\\
\textbf{Q1}: How does degree of similarity between feature spaces of source and target datasets impact performance when using TL?
\\
\textbf{Q2}: When a user is computationally / storage-constrained to pick only a small pretrained model, how would this affect performance of the resulting target model?
\\
\textbf{Q3}: Does TL alone help reduce ASR of a MIA on overfitted DNNs?

To address \textbf{Q1}, we consider two target datasets- CIFAR-10~\cite{krizhevsky2009CIFARs} and GTSRB~\cite{GTSRB013}- to learn a target model from a pretrained model that has been trained on ImageNet~\cite{deng2009imagenet} as the source dataset. These target datasets have different levels of similarity in their features with those of the source dataset. For example, the CIFAR-10 dataset has 10 classes, of which 9 are present in the ImageNet dataset. On the other hand, GTSRB has 43 classes of traffic signs; ImageNet has exactly one class for traffic signs. 
We also examine the role of shared features on a more complex face recognition task using CelebA~\cite{liu2015faceattributes} to learn a target model from a pretrained model trained on VGGFace2~\cite{cao2018vggface2} as source dataset. 
We present results of our experiments on these target datasets in \textbf{Sec. 6.1}. 

For \textbf{Q2} we examine three different SOTA pretrained models trained on ImageNet- VGG-19~\cite{simonyan2015very}, ResNet-18~\cite{he2016deep}, and Swin-T~\cite{liu2021swin}. 
The VGG-19 model consists of a DNN with $38$ layers; ResNet-18 consists of DNNs with $62$ layers. 
ResNet uses `shortcut' connections to overcome a \emph{vanishing gradient problem}~\cite{he2016deep} that is known to occur in large, complex DNN models during training. 
Swin-T uses a hierarchical transformer architecture and consists of $12$ layers distributed in $4$ blocks. 
Results of our experiments on VGG-19, ResNet-18, and Swin-T are presented in \textbf{Sec. 6.2}. 

To answer 
\textbf{Q3}, we compare the performance of Double-Dip Stage-1 with SOTA training-phase defenses against MIAs, including regularization~\cite{nasr2018machine} and distillation training~\cite{tang2022mitigating}. 
Regularization (e.g., $L1$-, $L2$-regularization) has been shown to mitigate overfitting~\cite{nasr2018machine}. 
However, these methods have been known to destroy trained features, thereby reducing classification accuracy of the target model for nonmembers~\cite{nasr2018machine, rajabi2022ldl, shokri2017membership}. 
Distillation-based training techniques were used in~\cite{tang2022mitigating}, which proposed SELENA, a framework which partitioned a given training set into subsets, and used these subsets to train `submodels', which were combined in an ensemble manner to train a final distilled model. 
Since target datasets for overfitted DNNs are small, submodels of SELENA might be prone to overfitting, consequently reducing classification accuracy of the final distilled model. 
We verify this hypothesis 
in \textbf{Sec. 6.3}. 

\subsection{Stage-2 of Double-Dip}
While the use of transfer learning in Stage-1 yields a target model embedded in a higher dimensional space that is less overfitted, thus readily reducing the success rate of an adversary carrying out a MIA\cite{jia2019memguard, rajabi2022ldl, shokri2017membership}, we are interested in reducing it further. 
Stage-2 employs a lightweight post-processing module that 
seeks to further reduce the ASR value of label-only MIAs without needing to retrain target models. 
Using terminology from \cite{cohen2019certified}, we denote a classifier $f$ with randomization as a \emph{smoothed classifier} $g$. 
For an input sample $x$, the smoothed classifier $g$ returns the same output label for all samples $x'$ inside a high-dimensional sphere of radius $r$ centered at $x$. 
A given sample $x$ is perturbed by a zero-mean Gaussian noise with variance $\sigma^2$. 
Stage-2 of Double-Dip 
tunes the value of 
of $\sigma$ to lower the ASR while maintaining high classification accuracy. 
We hypothesize that using Stages-1 \& 2 together will result in a lower ASR compared to using Stage-1 alone. 
We perform extensive experiments to verify this hypothesis in \textbf{Sec. 6.4}. 
Our experiments in \textbf{Sec. 6.4} also compare randomization-based approach of Stage 2 with regularization and differential privacy-based mechanisms.\\

\noindent \textbf{User Capability: }
For Stage-1, the user is assumed to have access to a limited number of samples (target dataset) to train a DNN to accomplish a desired classification task. 
The user also has access to publicly available pretrained models and code libraries. 
In Stage-2, the user has access to a mechanism to generate zero-mean Gaussian noise with given variance that can be added to a given sample to generate a high-dimensional region of constant output label centered at the sample. 

\subsection{Double-Dip Algorithm}
We describe the working of Double-Dip in Algorithm \ref{alg:DoubleDip}.

\begin{algorithm}
\caption{Double-Dip}\label{alg:DoubleDip}
\begin{algorithmic}[1]
\Require pretrained DNN model with $K$ layers; $N$ samples of target dataset $D_t$; $M$ frozen layers of pre-trained DNN
\State \textbf{Stage 1}: 
\State initialize target model weights to values of weights of pretrained model
\State freeze (fix) weights in $M$ layers of target model
\State use $N$ samples from $D_t$ to update weights of remaining $(K-M)$ layers using stochastic gradient descent
\State (\textbf{Stage 1 output}) Learned target model, ASR, Classification accuracy 

\State \textbf{Stage 2}: 
\State $n_i \sim \mathcal{N}(0,\sigma^2I)$; $x_{noisy}\leftarrow x + n_i$, $x \in D_t$, $i=1,2,\dots, s$
\State provide $x_{noisy}$ to learned target model from \textbf{Stage 1}
\State average predictions such that for $x \in D_t$, all $x_{noisy}$ result in same output label
\State (\textbf{Stage 2 output}) Protected target model, ASR, Classification accuracy 
\end{algorithmic}
\end{algorithm} 
\section{Experiment Settings}\label{sec:Evaluation}

\begin{table*}[!h]
\centering
\caption{\textbf{Stage-1 of Double-Dip, Pretrained VGG-19 Model}: Adversary success rate (ASR, \emph{lower is better}) and classification accuracy (ACC, \emph{higher is better}) for CIFAR-10 and GTSRB datasets with training sets of sizes 500 and 1000. 
We compare (i) no transfer learning (NTL), (ii) regularization (L1, L2), and (iii) transfer learning (TL). TL-X indicates that X layers of the pretrained model are frozen. 
We examine scenarios when an adversary carrying out an MIA has (a) white-box model access (BIM), and (b) black-box model access (HSJ). 
The best ASR and ACC values for a given training set size across both datasets is in \textbf{bold}; best ASR and ACC values in each cell are \underline{underlined}. 
TL yields lowest ASR values while also ensuring significantly higher accuracy. 
(See \textbf{Sec. 6.1}.)
}\label{tab:VGG-19-Target}
    \begin{tabular}{c|c|ccc|ccc}
        \toprule \hline
       \multicolumn{2}{c}{} & \multicolumn{3}{c}{\textbf{500}} & \multicolumn{3}{c}{\textbf{1000}}\\
        \hline
        \textbf{Dataset} & \textbf{Setting} & \%\textbf{ASR(BIM)} & \%\textbf{ASR(HSJ)} & \%\textbf{ACC} & \%\textbf{ASR(BIM)} & \%\textbf{ASR(HSJ)}& \%\textbf{ACC} \\
        \hline \hline
        \multirow{6}{*}{\thead{CIFAR-10}} & NTL & 87.5 & 87.5 &  24.6 & 88.7 & 88.5 & 27.7 \\
        & L1(0.001) & 90.1  & 88.9  & 23.6  &  86.5 & 85.8 & 28.0\\
        & L2(0.1) & 89.7   & 88.9 & 23.0 & 83.8 & 84.9 & 30.3 \\
        & \textbf{TL-0} & \cellcolor{gray!25} 60.1 & \cellcolor{gray!25} 60.6 & \cellcolor{gray!25} \underline{\textbf{79.2}}  & \cellcolor{gray!25} 59.9 & \cellcolor{gray!25} \underline{61.5} &  \cellcolor{gray!25} \underline{80.9} \\
        & \textbf{TL-20} & \cellcolor{gray!25} \underline{\textbf{59.9}} & \cellcolor{gray!25} \underline{\textbf{60.3}} & \cellcolor{gray!25} 78.6 & \cellcolor{gray!25} \underline{59.4} & \cellcolor{gray!25} \underline{61.5} &  \cellcolor{gray!25} 80.0 \\
        & \textbf{TL-35} & \cellcolor{gray!25} 62.9 & \cellcolor{gray!25} 63.5 & \cellcolor{gray!25} 72.2 & \cellcolor{gray!25} 63.7 &   \cellcolor{gray!25} 63.9 &  \cellcolor{gray!25} 76.1 \\
         \hline \hline 
        \multirow{6}{*}{\thead{GTSRB}} & NTL &76.0 &   76.7 &  40.8 &  76.7 & 74.8 & 54.3  \\
        & L1 (0.001) & 82.0 &81.5& 37.2 & 69.7 &   69.5 & 61.9 \\
        & L2 (0.1) &76.2 & 76.2& 43.8 &67.8&67.3& 62.9 \\
        & \textbf{TL-0} & \cellcolor{gray!25} \underline{63.0} & \cellcolor{gray!25} \underline{63.0} & \cellcolor{gray!25} 73.2 & \cellcolor{gray!25} \underline{\textbf{58.7}} &  \cellcolor{gray!25} \underline{\textbf{57.0}} &  \cellcolor{gray!25}\underline{\textbf{85.9}} \\
        & \textbf{TL-20} & \cellcolor{gray!25} \underline{63.0} & \cellcolor{gray!25} \underline{63.0} & \cellcolor{gray!25} \underline{73.6} &  \cellcolor{gray!25} 61.3 &  \cellcolor{gray!25} 62.0 &  \cellcolor{gray!25} 81.5 \\
        & \textbf{TL-35} & \cellcolor{gray!25} 70.0 & \cellcolor{gray!25} 70.0 & \cellcolor{gray!25} 59.0 & \cellcolor{gray!25}67.3  & \cellcolor{gray!25}68.8 &  \cellcolor{gray!25} 64.4\\
        \hline \bottomrule
    \end{tabular}
\end{table*}

We use two adversarial learning methods~\cite{chen2020hopskipjumpattack, BIM}. 
When the adversary has \emph{white-box model access}, it uses the BIM~\cite{BIM} to estimate the minimum noise needed to be added to a sample so that the sample is misclassified by the DNN. 
The BIM is a SOTA computationally inexpensive algorithm involving repeated application of the FGS method~\cite{goodfellow2014FGS}, where FGS generates adversarial perturbations 
in the direction of the gradient of a loss function. 
When the adversary has \emph{black-box model access}, it uses a SOTA query-efficient adversarial learning technique HopSkipJump~\cite{chen2020hopskipjumpattack} to learn the minimum noise required to cause the DNN to misclassify a given sample. 

We mainly use two target datasets- CIFAR-10~\cite{krizhevsky2009CIFARs} and GTSRB~\cite{GTSRB013} and training sets of sizes $ 500, 1000$. 
We use VGG-19~\cite{simonyan2015very}, ResNet-18~\cite{he2016deep}, and Swin-T~\cite{liu2021swin} models trained on ImageNet~\cite{deng2009imagenet} (from the Torchvision library~\cite{marcel2010torchvision}) as the pretrained model 
to learn each target model. 
The number of epochs for which a model is trained is chosen based on the maximum training accuracy achieved- for e.g., we choose 400 training epochs for VGG-19 and ResNet-18, and 900 training epochs for Swin-T. 
For VGG-19, we evaluate impacts of `freezing' $0, 20,$ and $35$ layers on ASR and classification accuracy of an adversary carrying out label-only MIAs~\cite{choquette2021label}. 
For ResNet-18, we `freeze' $0, 20, 40, 50,$ and $60$ layers. 
For Swin-T, we `freeze' $2$ layers. 
To evaluate performance of Double-Dip on a complex task like face recognition we perform experiments using FaceNet. For FaceNet~\cite{schroff2015facenet}, pretrained on VGGFace2~\cite{cao2018vggface2}, we evaluate the effect of freezing the first 300 layers for transfer learning with CelebA~\cite{liu2015faceattributes} as the target dataset.
For experiments `without transfer learning', we use randomly initialized weights of pretrained models and no frozen layers. 
We also compare Double-Dip with SOTA training-phase defenses against MIAs, including regularization~\cite{nasr2018machine}, distillation training~\cite{tang2022mitigating}, and differential privacy \cite{abadi2016deep}.

\section{Evaluation}\label{sec:evaluation}
This section presents results of our experiments. We use the metrics from Sec. \ref{sec:preliminaries} to evaluate the performance of \textbf{Double-Dip}. 
We will release code with the final version of the paper. 
We first evaluate Double-Dip Stage-1 by examining the effectiveness of transfer learning. 
We consider two cases: when the adversary has (i) white-box model access, and (ii) black-box model access. 
In each case, the adversary carries out a label-only MIA to estimate a threshold $\delta$ that will result in a given sample being misclassified by the target model. 
We then evaluate Double-Dip Stage-2 to investigate if ASR can be reduced further, without reducing accuracy. 
\begin{table}[!h]
\caption{\textbf{Stage-1 of Double-Dip, Pretrained FaceNet Model}: Adversary success rate (ASR, \emph{lower is better}) and classification accuracy (ACC, \emph{higher is better}) for the CelebA dataset with training size 1000. 
We compare (i) no transfer learning (NTL), (ii) regularization (L1, L2), and (iii) transfer learning (TL) when $300$ layers of the pretrained model are frozen (TL-300). 
We examine scenarios when an adversary carrying out an MIA has (a) white-box (BIM), and (b) black-box (HSJ) model access. 
The best ASR/ ACC values are in \underline{underlined} \textbf{bold}. 
Transfer learning yields lowest ASR while also ensuring significantly higher classification accuracy. (See \textbf{Sec. 6.1}.)
}\label{tab:FaceNet-Target}
\centering
    \begin{tabular}{c|c|c|c}
        \toprule \hline
        \textbf{Setting} & \%\textbf{ASR(BIM)}  & \%\textbf{ASR(HSJ)}  & \%\textbf{ACC}\\
        \hline \hline
        NTL & 66.5  &  78.8 & 41.7  \\
        L1(0.001) & 66.9 & 78.4 & 34.5 \\
        L2(0.1) & 61.4 & 78.0 & 40.0 \\
        \textbf{TL-300} & \cellcolor{gray!25} \underline{\textbf{56.8}} & \cellcolor{gray!25} \underline{\textbf{55.1}} &  \cellcolor{gray!25} \underline{\textbf{88.1}}\\
        \hline \bottomrule
    \end{tabular}
\end{table}

\begin{table*}[!h]
\centering
\caption{\textbf{Stage-1 of Double-Dip, Pretrained ResNet-18 Model}: Adversary success rate (ASR, \emph{lower is better}) and classification accuracy (ACC, \emph{higher is better}) for CIFAR-10 and GTSRB with training sets of sizes 500 and 1000. 
We compare (i) no transfer learning (NTL), (ii) regularization (L1, L2), and (iii) transfer learning (TL). TL-X indicates that X layers of the pretrained model are frozen. 
We examine scenarios when an adversary carrying out an MIA has (a) white-box model access (BIM), and (b) black-box model access (HSJ). 
The best ASR and ACC values for a given training set size across both datasets is in \textbf{bold}; best ASR and ACC values in each cell are \underline{underlined}. 
Transfer learning yields lowest ASR while also ensuring significantly higher classification accuracy. 
 (See \textbf{Sec. 6.2}.)
}\label{tab:ResNet-18-Target}
    \begin{tabular}{c|c|ccc|ccc}
        \toprule \hline
        \multicolumn{2}{c}{} & \multicolumn{3}{c}{\textbf{500}} & \multicolumn{3}{c}{\textbf{1000}}\\
        \hline
        \textbf{Dataset} & \textbf{Setting} & \%\textbf{ASR(BIM)} & \%\textbf{ASR(HSJ)} & \%\textbf{ACC} & \%\textbf{ASR(BIM)} & \%\textbf{ASR(HSJ)}& \%\textbf{ACC} \\
        \hline \hline
        \multirow{8}{*}{\thead{CIFAR-10}} & NTL & 81.7 &  81.7 &  36.2 &  78.8 & 79.1 & 43.0  \\
        & L1(0.001) & 81.7  & 82.5  & 36.4 &  79.1 &79.1 & 41.5\\
        & L2(0.1) & 82.2  & 82.7 & 35.8 & 77.9 & 78.1& 44.1 \\
        & \textbf{TL-0} & \cellcolor{gray!25} 64.4& \cellcolor{gray!25} 65.9& \cellcolor{gray!25} 65.8 & \cellcolor{gray!25} 61.1 & \cellcolor{gray!25} 61.5&  \cellcolor{gray!25} 77.3 \\
        & \textbf{TL-20} & \cellcolor{gray!25} 60.3 & \cellcolor{gray!25} 61.1& \cellcolor{gray!25} 73.2 & \cellcolor{gray!25} 63.0 &  \cellcolor{gray!25}62.7&\cellcolor{gray!25} 73.9 \\
        & \textbf{TL-40} & \cellcolor{gray!25} 59.9 & \cellcolor{gray!25}61.3 & \cellcolor{gray!25} 76.2 & \cellcolor{gray!25} 62.5&  \cellcolor{gray!25} 62.7& \cellcolor{gray!25} 77.4 \\	
        & \textbf{TL-50} & \cellcolor{gray!25} \underline{\textbf{58.9}} &  \cellcolor{gray!25} \underline{59.9} &\cellcolor{gray!25} \underline{{77.8}} & \cellcolor{gray!25} \underline{59.9} & \cellcolor{gray!25} \underline{60.6} &  \cellcolor{gray!25} \underline{81.2}\\	
        & \textbf{TL-60} & \cellcolor{gray!25} 61.3 &\cellcolor{gray!25} 62.3 &  \cellcolor{gray!25} 72.6 & \cellcolor{gray!25} 64.7 &   \cellcolor{gray!25} 64.9 &\cellcolor{gray!25} 72.5 \\	
        
        \hline \hline 	
        \multirow{8}{*}{\thead{GTSRB}} & NTL &  77.2 & 77.2&   42.0 &  73.8 &70.7 & 61.4  \\
        & L1 (0.001) & 74.8 & 75.0 & 41.4 & 72.4 & 72.1& 58.9 \\
        & L2 (0.1) &78.6 & 76.7 & 44.2& 73.1&71.6 & 60.3 \\
        & \textbf{TL-0} & \cellcolor{gray!25} 70.2 & \cellcolor{gray!25} 70.2 & \cellcolor{gray!25} 60.6 & \cellcolor{gray!25} 57.5 & \cellcolor{gray!25}58.7&  \cellcolor{gray!25} 81.3\\
        & \textbf{TL-20} & \cellcolor{gray!25} 68.0 &  \cellcolor{gray!25} 68.0 & \cellcolor{gray!25} 65.8 & \cellcolor{gray!25} 62.5 &   \cellcolor{gray!25}64.9 &  \cellcolor{gray!25} 75.9 \\
        & \textbf{TL-40} & \cellcolor{gray!25} \underline{{56.0}} & \cellcolor{gray!25} \underline{\textbf{54.1}} & \cellcolor{gray!25} \underline{\textbf{89.2}}  & \cellcolor{gray!25} \underline{\textbf{54.8}} &  \cellcolor{gray!25} \underline{\textbf{53.8}} &  \cellcolor{gray!25} \underline{\textbf{92.3}}\\
        & \textbf{TL-50} & \cellcolor{gray!25} 61.1 & \cellcolor{gray!25} 61.1 & \cellcolor{gray!25} 78.2 & \cellcolor{gray!25}59.1 & \cellcolor{gray!25}58.9 &  \cellcolor{gray!25} 82.9\\
        & \textbf{TL-60} & \cellcolor{gray!25} 67.1 &  \cellcolor{gray!25} 67.1 & \cellcolor{gray!25} 64.2 & \cellcolor{gray!25}66.8& \cellcolor{gray!25}69.0&  \cellcolor{gray!25} 63.1\\
        \hline \bottomrule
    \end{tabular}
\end{table*}
%

\subsection{Role of Correlated Features}
Table \ref{tab:VGG-19-Target} compares ASR and accuracy when (i) using Stage-1 of Double-Dip on a \emph{pretrained VGG-19 model}, (ii) without transfer learning, and (iii) when using L1/ L2 regularization. 
We consider scenarios where an adversary carrying out a MIA has (a) white-box model access, and (b) black-box model access. 
We examine CIFAR-10 and GTSRB target datasets of sizes $500$ and $1000$. 
Our results reveal that for both datasets, \emph{transfer learning used in Stage-1 of Double-Dip is effective in reducing the ASR of an adversary carrying out a MIA while also achieving significantly higher classification accuracy} relative to no transfer learning or regularization. 

Further, for a smaller number of training samples, classification accuracies of nonmembers when using transfer learning is higher when the target dataset is CIFAR-10 than for GTSRB. 
This is because CIFAR-10 is highly correlated with ImageNet specifically, as labels of samples from $9$ out of $10$ classes in CIFAR-10 can be found in the ImageNet dataset~\cite{chrabaszcz2017downsampled}. 
On the other hand, GTSRB consists of samples belonging to one of $43$ classes of traffic signs; ImageNet has exactly one class for traffic signs. 
Increasing the number of training samples is somewhat effective in improving classification accuracy. 
We observe that Stage-1 of Double-Dip is effective in reducing ASR 
irrespective of whether an adversary carrying out a MIA has white-box or black-box model access. 

Table \ref{tab:FaceNet-Target} indicates that Stage-1 of Double-Dip is effective in reducing ASR while maintaining high classification accuracy when using the CelebA dataset to learn a target model using a \emph{pretrained FaceNet model}. 
This is due to the fact that the VGGFace2 source dataset \cite{cao2018vggface2} used to train the FaceNet model \cite{schroff2015facenet} has multiple features in common with the CelebA target dataset, which is a dataset of faces of celebrities with each image having $40$ attribute annotations \cite{liu2015faceattributes}. 

\begin{table*}[!h]
\centering
\caption{\textbf{Stage-1 of Double-Dip, Pretrained Swin Model}: Adversary success rate (ASR, \emph{lower is better}) and classification accuracy (ACC, \emph{higher is better}) for CIFAR-10 and GTSRB with training sets of sizes 500 and 1000. 
We compare (i) no transfer learning (NTL), (ii) regularization (L1, L2), and (iii) transfer learning when $2$ layers of the pretrained model are frozen (TL-2). 
We examine scenarios when an adversary carrying out an MIA has (a) white-box (BIM), and (b) black-box (HSJ) model access. 
Best ASR and ACC values for a given training set size across both datasets is in \textbf{bold}; best ASR and ACC values in each cell are \underline{underlined}. 
Transfer learning yields lowest ASR while also ensuring significantly higher classification accuracy. 
(See \textbf{Sec. 6.2}.)
}\label{tab:Swin-t-Target}
    \begin{tabular}{c|c|ccc|ccc}
        \toprule \hline
        \multicolumn{2}{c}{} & \multicolumn{3}{c}{\textbf{500}} & \multicolumn{3}{c}{\textbf{1000}}\\
        \hline
        \textbf{Dataset} & \textbf{Setting} & \%\textbf{ASR(BIM)} & \%\textbf{ASR(HSJ)} & \%\textbf{ACC} & \%\textbf{ASR(BIM)} & \%\textbf{ASR(HSJ)}& \%\textbf{ACC} \\
        \hline \hline	
        \multirow{4}{*}{\thead{CIFAR-10}} & NTL & 63.5 &  65.0 &  34.4 & 66.6 &  67.5 & 40.8 \\
        & L1(0.001) & 63.3  & 64.0 & 34.2 &  62.0 &64.0 &38.4 \\
        & L2(0.1) & 65.0  & 66.0  & 34.2 & 68.5 & 66.8 & 37.5 \\
        & \textbf{TL-2} & \cellcolor{gray!25} \underline{\textbf{54.6}}& \cellcolor{gray!25} \underline{\textbf{57.3}}& \cellcolor{gray!25} \underline{\textbf{89.2}}  & \cellcolor{gray!25} \underline{56.7} & \cellcolor{gray!25} \underline{\textbf{54.8}} &  \cellcolor{gray!25} \underline{88.6}\\	
        \hline \hline 	
        \multirow{4}{*}{\thead{GTSRB}} & NTL &62.0&75.0&40.8&58.8 & 69.0 & 63.6  \\
        & L1 (0.001) &59.3 &62.5 & 35.2 &54.8 &56.0 & 39.8\\
        & L2 (0.1) &59.8 &67.5 & 38.0 &63.8 & 72.3 & 53.6\\
        & \textbf{TL-2} & \cellcolor{gray!25} \underline{56.8}& \cellcolor{gray!25} \underline{60.5 }& \cellcolor{gray!25} \underline{78.8} & \cellcolor{gray!25} \underline{\textbf{52.5}} &  \cellcolor{gray!25} \underline{55.0} &  \cellcolor{gray!25} \underline{\textbf{91.3}}\\
        \hline \bottomrule
    \end{tabular}
\end{table*}

\begin{table}[h]
\centering 
\caption{\textbf{Stage-1 of Double-Dip vs. SOTA}: 
Adversary success rate ($ASR$) and classification accuracy of nonmembers ($ACC$) using $1000$ training samples each from CIFAR-10 and GTSRB. 
We compare (i) no transfer learning (NTL), (ii) a SOTA self-distillation defense for MIA, SELENA~\cite{tang2022mitigating}, and (iii) transfer learning (TL) using a \textbf{pretrained ResNet-18} model. TL-X indicates that X layers of the pretrained model are frozen. 
We examine scenarios when an adversary carrying out an MIA has (a) white-box model access (BIM), and (b) black-box model access (HSJ). 
For CIFAR-10, although SELENA achieves lowest ASR, it is accompanied by a significant drop in ACC. Transfer learning, on the other hand, achieves much higher classification accuracy with similar $ASR$ values.  
For GTSRB, transfer learning consistently achieves lower ASR and higher ACC values in all settings. (See \textbf{Sec. 6.3}.)
}\label{tab:SELENA-MAIN}
\begin{tabular}{c|c|c|c|c}
\toprule \hline
\textbf{Dataset}&\textbf{Setting} & \%\textbf{ASR(BIM)} &\% \textbf{ASR(HSJ)}&  \%\textbf{ACC}  \\
\hline \hline	
 \multirow{6}{*}{\thead{CIFAR-10}}&NTL         & 78.6& 78.8 &  $44.1$       \\ 
 &SELENA        &$\mathbf{\underline{55.0}}$ & $\mathbf{\underline{55.6}}$  &  $36.8$      \\ 
&\textbf{TL-0}   & \cellcolor{gray!25} 61.5 & \cellcolor{gray!25} 61.8  & \cellcolor{gray!25} $\underline{76.9}$      \\ 
 &\textbf{TL-20} & \cellcolor{gray!25} 62.3 & \cellcolor{gray!25} 62.5 & \cellcolor{gray!25} 74.7\\ 
 &\textbf{TL-50}      & \cellcolor{gray!25} 62.7 &\cellcolor{gray!25}  63.9 &  \cellcolor{gray!25} 71.7 \\ 
 &\textbf{TL-60}      & \cellcolor{gray!25} 65.4 & \cellcolor{gray!25} 64.9 & \cellcolor{gray!25} 70.5  \\ 
 \hline \hline
 \multirow{6}{*}{\thead{GTSRB}} &NTL            & 72.6 & 70.9 & {$61.3$}     \\ 
 &SELENA           & 82.0 & 83.8 &   ${31.2}$      \\ 
&\textbf{TL-0}   &\cellcolor{gray!25}  $\underline{57.2}$ & $\cellcolor{gray!25} \underline{58.9}$  &   $\cellcolor{gray!25} \mathbf{\underline{81.9}}$    \\ 
 &\textbf{TL-20}      &\cellcolor{gray!25}  62.7 & \cellcolor{gray!25} 64.9  &   $\cellcolor{gray!25} 76.6$  \\ 
 &\textbf{TL-50}      &\cellcolor{gray!25}  67.5  & \cellcolor{gray!25} 67.5 &   $\cellcolor{gray!25} 66.4 $ \\ 
 &\textbf{TL-60}     &\cellcolor{gray!25}  66.6  &\cellcolor{gray!25}  69.0 &   $\cellcolor{gray!25} 63.6$ \\ 
 \hline \bottomrule
\end{tabular}
\end{table}

\subsection{Complexity of Pretrained Models}
We compare ASR and classification accuracy of Stage-1 of Double-Dip when using a \emph{pretrained VGG-19 model} in Table \ref{tab:VGG-19-Target} and two more complex models- a \emph{pretrained ResNet-18 model} in Table \ref{tab:ResNet-18-Target}, and a transformer-based \emph{pretrained Swin-T model} in Table \ref{tab:Swin-t-Target} for the CIFAR-10 and GTSRB target datasets of sizes $500$ and $1000$. 
When fewer layers are frozen (compare \textbf{TL-0} in Tables \ref{tab:VGG-19-Target} and \ref{tab:ResNet-18-Target}), a less complex model (e.g., VGG-19) is more effective in reducing ASR while achieving high classification accuracy. 
This is because a larger number of model parameters will have to be learned for ResNet-18 compared to VGG-19 for the same size of target dataset. 
However, as the number of frozen layers for transfer learning increases, the complexity of models such as ResNet-18 and Swin-T suggest greater flexibility in reducing the ASR while also accomplishing higher classification accuracy for nonmembers. 

%
\subsection{Stage-1 of Double-Dip vs. SOTA}
Table \ref{tab:SELENA-MAIN} compares Stage-1 of Double-Dip using a \emph{pretrained ResNet-18 model} with SELENA~\cite{tang2022mitigating}, a SOTA self-distillation defense against MIAs. We use $1000$ training samples from CIFAR-10 and GTSRB as target datasets. 
We consider two types of label-only MIAs by an adversary with: (i) black-box model access using HopSkipJump~\cite{chen2020hopskipjumpattack}, and (ii) white-box model access using BIM~\cite{BIM}. 
For CIFAR-10, although SELENA achieves lowest ASR, it is accompanied by significant reduction in classification accuracy. 
An explanation 
is that SELENA partitions the training dataset into subsets to train submodels, and then uses an ensemble of submodels to train the final model~\cite{tang2022mitigating}. 
However, when the target dataset is small (e.g., our case of $1000$ samples from CIFAR-10), submodels learned by SELENA can be overfitted, which affects classification accuracy, thereby rendering the learned model less useful to a user. 
In comparison, using transfer learning achieves similar ASR values with $\mathbf{>2}$\textbf{x} higher accuracy. 
For GTSRB, transfer learning consistently achieves lower ASR and $\mathbf{>2}$\textbf{x} higher classification accuracy than SELENA. 
%
\begin{figure}[!h]
    \centering
    \begin{tabular}{c}
    \includegraphics[trim={1cm 0cm 0.5cm 0cm }, scale=0.18]{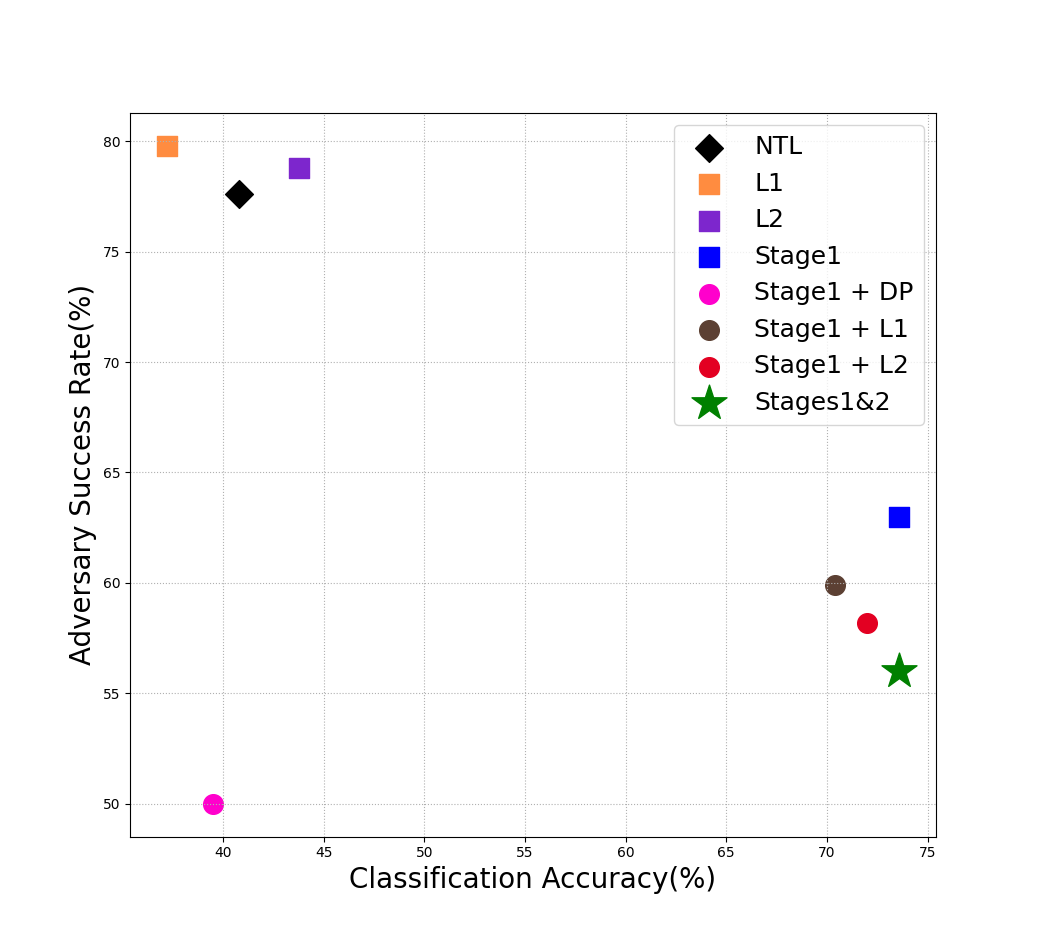}
    \end{tabular}
    \caption{\textbf{Stages-1\&2 of Double-Dip vs. SOTA}: Adversary success rate (ASR, \emph{lower is better}) and classification accuracy (ACC, \emph{higher is better}) for 500 training samples from GTSRB with a pretrained VGG-19 model when using (i) no transfer learning (NTL), (ii) regularization (L1/ L2), (iii) Stage-1 of Double-Dip, (iv) Double-Dip Stage-1 + diff. privacy (Stage-1+DP), (v) Stage-1 of Double-Dip + regularization (Stage-1+L1/ L2), and (vi) Stages-1\&2 of Double-Dip. Stages-1\&2 of Double-Dip achieves low ASR values while simultaneously ensuring high ACC. While Stage-1+DP achieves lowest ASR, it comes at the cost of a significant reduction in accuracy. 
    }
    \label{fig:DoubleDefenseComparison}
\end{figure}
\begin{figure}
    \centering
    \begin{tabular}{c}
    \includegraphics[trim={1cm 0cm 0.5cm 0cm }, scale=0.2]{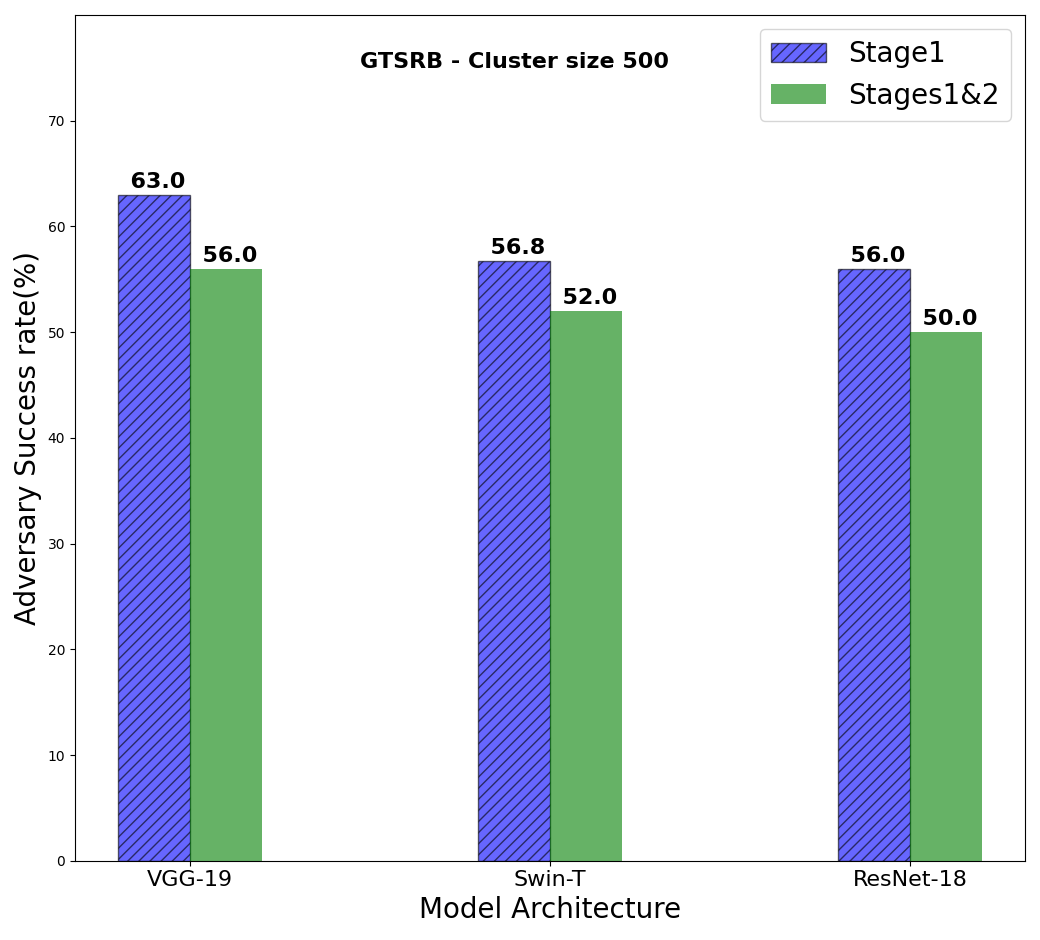}
    \\
    \includegraphics[trim={1cm 0cm 0.5cm 0cm }, scale=0.2]{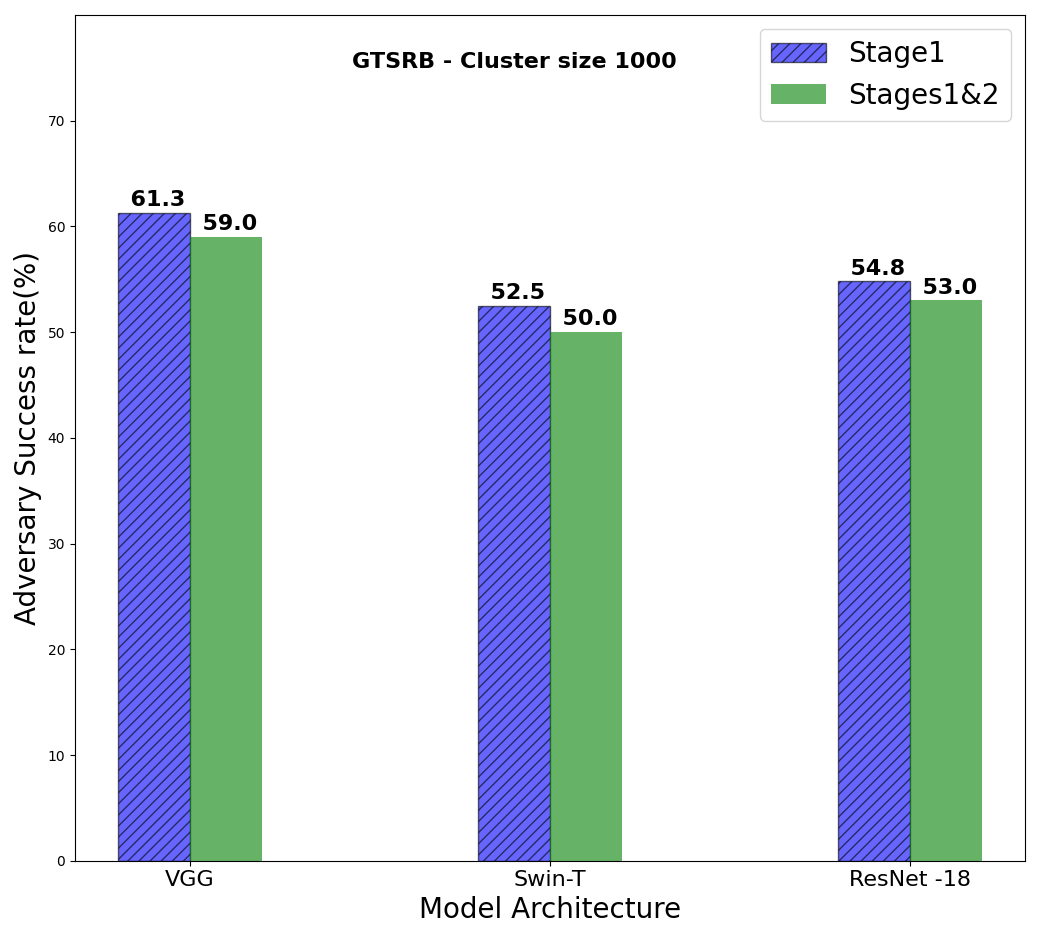}
    \end{tabular}
    \caption{\textbf{Stages-1\&2 of Double-Dip Reduces ASR}: Comparison of ASR values when using Stage-1 of Double-Dip (blue bars) and Stages-1\&2 of Double-Dip (green bars) for $500$ (top) and $1000$ (bottom) training samples from the GTSRB dataset when an adversary carrying out a MIA has white-box model access. Stages-1\&2 of Double-Dip is effective in reducing the ASR value, relative to Stage-1 of Double-Dip when using three different types of pretrained models- VGG-19, ResNet-18, and Swin-T.
    }
    \label{fig:Stage2Comparison}
\end{figure}
\subsection{Stage-2 of Double-Dip: Reducing ASR}
Although transfer learning alone was effective in reducing the ASR, we would like to reduce it further. 
An adversary carrying out a label-only MIA uses the magnitude of noise perturbations required to misclassify a given sample to determine if it is a member (large noise) or not (small noise)~\cite{choquette2021label, rajabi2022ldl}. 
Creating ambiguity between members and nonmembers will lower the ASR. 
Towards this, we use a smoothed classifier~\cite{cohen2019certified} to add calibrated Gaussian noise of given variance to data samples (\textbf{Stage-2} in Sec. \ref{sec:methodologyDoubleDip}). 
Such a process lowers ASR by ensuring that magnitudes of noise perturbations required to misclassify members and nonmembers will be comparable. 

Fig. \ref{fig:DoubleDefenseComparison} compares ASR and accuracy when using different SOTA mechanisms with the goal of reducing ASR relative to ASR obtained in Stage-1 of Double-Dip. 
We show results for $500$ training samples from GTSRB with a pretrained VGG-19 model, and consider an adversary with white-box model access carrying out a MIA. 
We compare effects of using (i) regularization (L1/ L2), (ii) differential privacy, and (iii) randomization based on noise perturbation (Stage-2 of Double-Dip) on values of ASR and accuracy. 
A `better' method corresponds to a location in the \emph{bottom right corner} of 
Fig. \ref{fig:DoubleDefenseComparison}. 

\begin{figure}[!h]
    \centering
    \begin{tabular}{c}
    \includegraphics[trim={1cm 0cm 0.5cm 0cm }, scale=0.15]{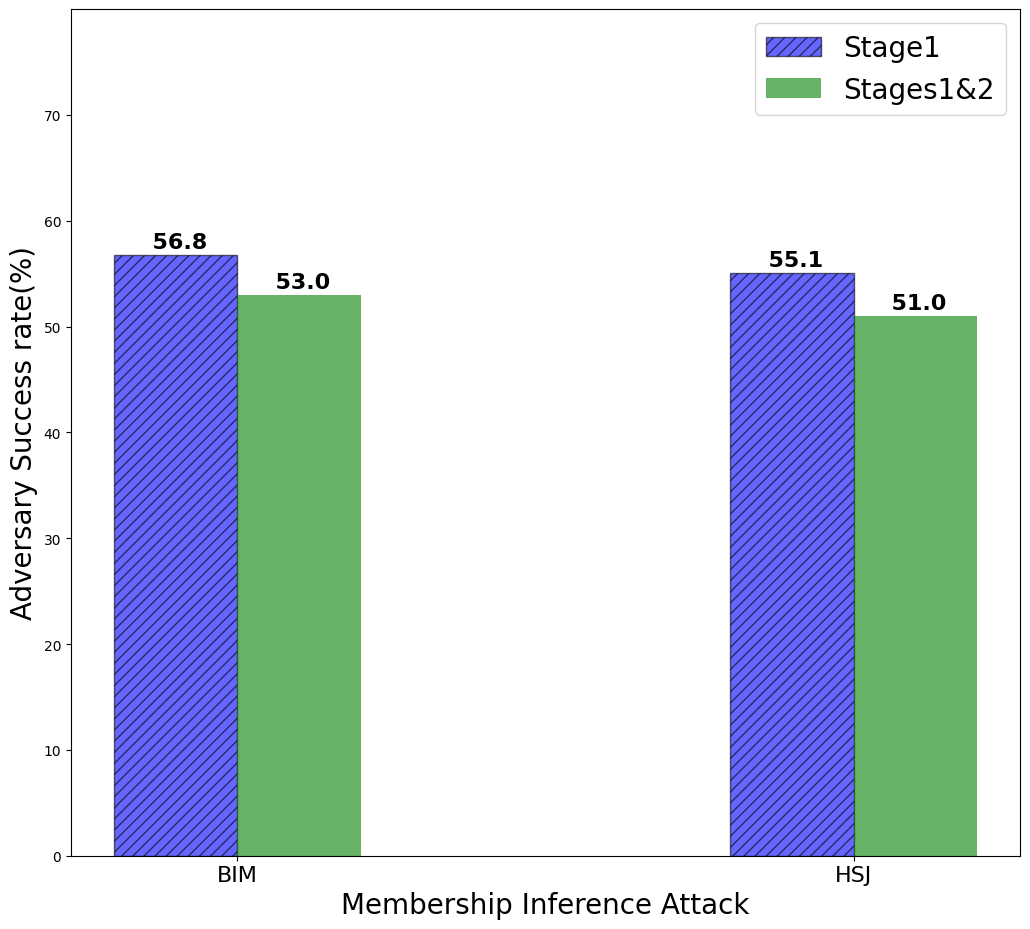}
    \end{tabular}
    \caption{\textbf{Stages-1\&2 of Double-Dip, Black-box Access}: Comparison of ASR values when using Stage-1 of Double-Dip (blue bars) and Stages-1\&2 of Double-Dip(green bars) for training samples from the CelebA dataset when an adversary carrying out a MIA has white-box (BIM, left) or black-box (HSJ, right) model access. We observe that Stages-1\&2 of Double-Dip is effective in reducing ASR in both cases. (\emph{Please zoom in for clarity})}
    \label{fig:FaceNet_Stage2Comparison}
\end{figure}

Our results show that Double-Dip Stages-1\&2 achieves low ASR while simultaneously ensuring high classification accuracy. 
Moreover, ASR in this case is lower than ASR obtained when using only Stage-1 of Double-Dip. 
Regularization-based techniques are effective in reducing ASR; however, they also come at the cost of reducing classification accuracy. 
Although differential privacy-based mechanisms achieve lowest ASR, this is accompanied by significant reduction in classification accuracy, which will impact usability.

Fig. \ref{fig:Stage2Comparison} shows that Stages-1\&2 of Double-Dip is effective in reducing the ASR when three pretrained models- VGG-19, ResNet-18, and Swin-T- are used to learn a target model for the GTSRB dataset with $500$ and $1000$ training samples. 
Fig. \ref{fig:FaceNet_Stage2Comparison} indicates similar effectiveness of Double-Dip Stages-1\&2 in reducing ASR when using CelebA as target dataset on a pretrained FaceNet model when an adversary carrying out a MIA has white or black-box access.

\begin{table*}[!h]
\centering
\caption{\textbf{Stage-1 of Double-Dip, Pretrained VGG-19 Shadow Model}: Adversary success rate (ASR, \emph{lower is better}) and classification accuracy (ACC, \emph{higher is better}) for CIFAR-10 and GTSRB datasets with training sets of sizes 500 and 1000. 
We compare (i) no transfer learning (NTL), (ii) regularization (L1, L2), and (iii) transfer learning (TL). TL-X indicates that X layers of the pretrained model are frozen. 
We examine scenarios when an adversary carrying out an MIA has (a) white-box model access (BIM), and (b) black-box model access (HSJ). 
The best ASR and ACC values for a given training set size across both datasets is in \textbf{bold}; best ASR and ACC values in each cell are \underline{underlined}. 
Transfer learning yields lowest ASR values while also ensuring significantly higher classification accuracy. 
(See \textbf{Sec. 6.5}.)
}\label{tab:VGG-19-Shadow}
    \begin{tabular}{c|c|ccc|ccc}
        \toprule \hline
       \multicolumn{2}{c}{} & \multicolumn{3}{c}{\textbf{500}} & \multicolumn{3}{c}{\textbf{1000}}\\
        \hline
        \textbf{Dataset} & \textbf{Setting} & \%\textbf{ASR(BIM)} & \%\textbf{ASR(HSJ)} & \%\textbf{ACC} & \%\textbf{ASR(BIM)} & \%\textbf{ASR(HSJ)}& \%\textbf{ACC} \\
        \hline \hline
        \multirow{6}{*}{\thead{CIFAR-10}} & NTL & 88.0 & 86.3 &  27.2 & 88.0 & 84.0 & 28.5 \\
        & L1(0.001) & 90.8  & 89.0  & 23.6  &  86.5 & 85.3 & 25.1\\
        & L2(0.1) & 89.8  & 89.3 & 21.2 & 84.8 & 84.8 & 24.1 \\
        & \textbf{TL-0} & \cellcolor{gray!25} \underline{\textbf{59.5}} & \cellcolor{gray!25} 61.3 & \cellcolor{gray!25} \underline{74.8}  & \cellcolor{gray!25} 61.3 & \cellcolor{gray!25} 61.0 &  \cellcolor{gray!25} \underline{81.2} \\
        & \textbf{TL-20} & \cellcolor{gray!25} 61.3 & \cellcolor{gray!25} \underline{\textbf{60.7}} & \cellcolor{gray!25} 72.6 & \cellcolor{gray!25} \underline{\textbf{60.3}} & \cellcolor{gray!25} \underline{60.5} &  \cellcolor{gray!25} 80.7 \\
        & \textbf{TL-35} & \cellcolor{gray!25} 63.8 & \cellcolor{gray!25} 64.0 & \cellcolor{gray!25} 73.0 & \cellcolor{gray!25} 64.8 &   \cellcolor{gray!25} 63.5 &  \cellcolor{gray!25} 78.0 \\
         \hline \hline 
        \multirow{6}{*}{\thead{GTSRB}} & NTL &75.3 &   77.8 &  46.0 &  76.8 &75.0& 54.5  \\
        & L1 (0.001) & 82.3 &81.0& 36.0 & 69.3 & 69.3 & 59.4 \\
        & L2 (0.1) &77.0 & 76.5& 38.4 &68.0&68.3& 56.2 \\
        & \textbf{TL-0} & \cellcolor{gray!25} \underline{62.3} & \cellcolor{gray!25} \underline{63.8} & \cellcolor{gray!25} \underline{\textbf{75.0}} & \cellcolor{gray!25} \underline{60.5} &  \cellcolor{gray!25} \underline{\textbf{58.8}} &  \cellcolor{gray!25}\underline{\textbf{87.2}} \\
        & \textbf{TL-20} & \cellcolor{gray!25} 63.0 & \cellcolor{gray!25} 64.3 & \cellcolor{gray!25} 70.4 &  \cellcolor{gray!25} 60.7 &  \cellcolor{gray!25} 62.2 &  \cellcolor{gray!25} 83.3 \\
        & \textbf{TL-35} & \cellcolor{gray!25} 68.3 & \cellcolor{gray!25} 71.3 & \cellcolor{gray!25} 54.6 & \cellcolor{gray!25}68.8  & \cellcolor{gray!25}68.8 &  \cellcolor{gray!25} 66.5\\
        \hline \bottomrule
    \end{tabular}
\end{table*}

\begin{table*}[!h]
\centering
\caption{\textbf{Stage-1 of Double-Dip, Pretrained ResNet-18 Shadow Model}: Adversary success rate (ASR, \emph{lower is better}) and classification accuracy (ACC, \emph{higher is better}) for CIFAR-10 and GTSRB with training sets of sizes 500 and 1000. 
We compare (i) no transfer learning (NTL), (ii) regularization (L1, L2), and (iii) transfer learning (TL). 
We examine scenarios when an adversary carrying out an MIA has (a) white-box model access (BIM), and (b) black-box model access (HSJ). 
The best ASR and ACC values for a given training set size across both datasets is in \textbf{bold}; best ASR and ACC values in each cell are \underline{underlined}. 
Transfer learning yields lowest ASR while also ensuring significantly higher classification accuracy.
 (See \textbf{Sec. 6.5}.)
}\label{tab:ResNet-18-Shadow}
    \begin{tabular}{c|c|ccc|ccc}
        \toprule \hline
        \multicolumn{2}{c}{} & \multicolumn{3}{c}{\textbf{500}} & \multicolumn{3}{c}{\textbf{1000}}\\
        \hline
        \textbf{Dataset} & \textbf{Setting} & \%\textbf{ASR(BIM)} & \%\textbf{ASR(HSJ)} & \%\textbf{ACC} & \%\textbf{ASR(BIM)} & \%\textbf{ASR(HSJ)}& \%\textbf{ACC} \\
        \hline \hline
        \multirow{8}{*}{\thead{CIFAR-10}} & NTL & 81.3 &  80.5 &  38.2 &  78.8 & 77.1 & 39.0  \\
        & L1(0.001) & 82.5 & 81.5 & 36.8 & 78.8 & 79.1 & 40.9\\
        & L2(0.1) &  82.7 & 81.3 & 33.8 & 77.9 & 78.1 & 38.1  \\
        & \textbf{TL-0} & \cellcolor{gray!25} 65.4 & \cellcolor{gray!25} 65.9& \cellcolor{gray!25} 64.2 & \cellcolor{gray!25} 61.1 & \cellcolor{gray!25} 61.3&  \cellcolor{gray!25} 73.6 \\
        & \textbf{TL-20} & \cellcolor{gray!25} 60.3 & \cellcolor{gray!25} 62.5& \cellcolor{gray!25} 69.8 & \cellcolor{gray!25} 63.5 &  \cellcolor{gray!25}63.5 &\cellcolor{gray!25} 76.5 \\
        & \textbf{TL-40} & \cellcolor{gray!25} \underline{\textbf{52.4}} & \cellcolor{gray!25}61.5 & \cellcolor{gray!25} 72.6 & \cellcolor{gray!25} 62.5&  \cellcolor{gray!25} 63.2& \cellcolor{gray!25} 78.7 \\	
        & \textbf{TL-50} & \cellcolor{gray!25} 59.4 &  \cellcolor{gray!25} \underline{60.6} &\cellcolor{gray!25} \underline{{74.8}} & \cellcolor{gray!25} \underline{59.9} & \cellcolor{gray!25} \underline{59.9} &  \cellcolor{gray!25} \underline{78.8}\\	
        & \textbf{TL-60} & \cellcolor{gray!25} 61.1 &\cellcolor{gray!25} 62.3 &  \cellcolor{gray!25} 72.8 & \cellcolor{gray!25} 65.1 &   \cellcolor{gray!25} 65.1 &\cellcolor{gray!25} 69.6 \\	
        
        \hline \hline 	
        \multirow{8}{*}{\thead{GTSRB}} & NTL &  78.6 & 76.7 & 38.2 & 72.1 & 70.9 & 61.7  \\
        & L1 (0.001) & 75.0 & 75.2 & 36.4 & 72.6 & 72.1 & 60.7 \\
        & L2 (0.1) & 78.6 & 76.7 & 40.8 & 74.8 & 71.6 & 60.6 \\
        & \textbf{TL-0} & \cellcolor{gray!25} 70.0 & \cellcolor{gray!25} 70.4 & \cellcolor{gray!25} 67.6 & \cellcolor{gray!25} 56.5 & \cellcolor{gray!25}58.7&  \cellcolor{gray!25} 78.9\\
        & \textbf{TL-20} & \cellcolor{gray!25} 67.8 &  \cellcolor{gray!25} 68.3 & \cellcolor{gray!25} 61.2 & \cellcolor{gray!25} 64.4 &   \cellcolor{gray!25} 64.9 &  \cellcolor{gray!25} 78.2 \\
        & \textbf{TL-40} & \cellcolor{gray!25} \underline{{57.7}} & \cellcolor{gray!25} \underline{\textbf{54.8}} & \cellcolor{gray!25} \underline{\textbf{88.4}}  & \cellcolor{gray!25} \underline{\textbf{53.6}} &  \cellcolor{gray!25} \underline{\textbf{54.8}} &  \cellcolor{gray!25} \underline{\textbf{94.6}}\\
        & \textbf{TL-50} & \cellcolor{gray!25} 60.6 & \cellcolor{gray!25} 60.6 & \cellcolor{gray!25} 75.6 & \cellcolor{gray!25}58.2 & \cellcolor{gray!25}59.4 &  \cellcolor{gray!25} 86.3\\
        & \textbf{TL-60} & \cellcolor{gray!25} 65.6 &  \cellcolor{gray!25} 67.3 & \cellcolor{gray!25} 60.8 & \cellcolor{gray!25}68.8& \cellcolor{gray!25}69.2&  \cellcolor{gray!25} 69.1\\
        \hline \bottomrule
    \end{tabular}
\end{table*}

\begin{table}[h]
\centering 
\caption{\textbf{Stage-1 of Double-Dip vs. SOTA for Shadow Models}: 
This table compares adversary success rate ($ASR$) and classification accuracy of nonmembers ($ACC$) using $1000$ training samples each from CIFAR-10 and GTSRB 
when using (i) no transfer learning (NTL), (ii) a SOTA self-distillation defense for MIA, SELENA~\cite{tang2022mitigating}, and (iii) transfer learning using a \textbf{pretrained ResNet-18} model. TL-X indicates that X layers of the pretrained model are frozen. 
We examine scenarios when an adversary carrying out an MIA has (a) white-box access (BIM), and (b) black-box access (HSJ). 
For CIFAR-10, although SELENA achieves lowest ASR, it is accompanied by a significant reduction in classification accuracy. Transfer learning, on the other hand, achieves much higher classification accuracy with similar $ASR$ values.  
For GTSRB, transfer learning consistently achieves lower ASR and higher ACC values in all settings. (See \textbf{Sec. 6.5}.) 
}\label{tab:SELENA-MAIN-SHADOW}
\begin{tabular}{c|c|c|c|c}
\toprule \hline
\textbf{Dataset}&\textbf{Setting} & \%\textbf{ASR(BIM)} &\% \textbf{ASR(HSJ)}&  \%\textbf{ACC}  \\
\hline \hline	
 \multirow{6}{*}{\thead{CIFAR-10}}&NTL         & 78.6& 79.1&  $39.8$       \\ 
 &SELENA        &$\mathbf{\underline{54.6}}$ & $\mathbf{\underline{55.0}}$  &  $37.0$      \\ 
&\textbf{TL-0}   & \cellcolor{gray!25} 61.8 & \cellcolor{gray!25} 62.0  & \cellcolor{gray!25} $\underline{73.7}$      \\ 
 &\textbf{TL-20} & \cellcolor{gray!25} 62.7 & \cellcolor{gray!25} 62.5 & \cellcolor{gray!25} 77.0\\ 
 &\textbf{TL-50}      & \cellcolor{gray!25} 63.2 &\cellcolor{gray!25}  64.2 &  \cellcolor{gray!25} 74.4 \\ 
 &\textbf{TL-60}      & \cellcolor{gray!25} 65.9 & \cellcolor{gray!25} 65.9 & \cellcolor{gray!25} 67.9 \\ 
 \hline \hline
 \multirow{6}{*}{\thead{GTSRB}} &NTL            & 72.4 & 71.2 & {$61.5$}     \\ 
 &SELENA           & 71.8 & 78.3 &   ${62.5}$      \\ 
&\textbf{TL-0}   &\cellcolor{gray!25}  $\underline{56.3}$ & $\cellcolor{gray!25} \underline{58.9}$  &   $\cellcolor{gray!25} 79.2$    \\ 
 &\textbf{TL-20}      &\cellcolor{gray!25}  64.4 & \cellcolor{gray!25} 64.9  &   $\cellcolor{gray!25} \mathbf{\underline{79.6}}$  \\ 
 &\textbf{TL-50}      &\cellcolor{gray!25}  67.5  & \cellcolor{gray!25} 67.8 &   $\cellcolor{gray!25} 62.2 $ \\ 
 &\textbf{TL-60}     &\cellcolor{gray!25}  68.8  &\cellcolor{gray!25}  69.0 &   $\cellcolor{gray!25} 68.8$ \\ 
 \hline \bottomrule
\end{tabular}
\end{table}

\subsection{Use of Shadow Models}
An adversary may not have access to sufficient samples to allow it to estimate the threshold $\delta$ on the noise that needs to be added for the target DNN to misclassify a given sample. 
However, it might have access to datasets of other samples along with their labels. 
The authors of \cite{choquette2021label} proposed using such alternate datasets along with complete knowledge of parameters of the target DNN to train a shadow model (also termed substitute model in \cite{rajabi2022ldl}). 
The threshold $\delta$ was determined using the shadow model. 

Our results in Tables \ref{tab:VGG-19-Shadow} and \ref{tab:ResNet-18-Shadow} evaluating Stage-1 of Double-Dip indicate that shadow models learned by the adversary are a good representation of the target DNN. 
This is confirmed by the observation that Stage-1 of Double-Dip yields lowest ASR while also ensuring significantly higher classification accuracy when using pretrained VGG-19 and ResNet-18 models. 

We further compare the performance of Stage-1 of Double-Dip against SELENA \cite{tang2022mitigating}, a SOTA distillation-based defense in Table \ref{tab:SELENA-MAIN-SHADOW}. 
Our results reveal that although SELENA can be effective in reducing ASR values, it comes at the cost of a significant reduction in classification accuracy. 
On the other hand, Stage-1 of Double-Dip is effective in reducing ASR while simultaneously ensuring high classification accuracy. 
\\
Our results in this section show that \emph{Double-Dip} is an effective defense against MIAs for adversaries with either white-box or black-box access to the target model. Double-Dip ensures high classification accuracy for nonmembers while lowering ASR for label-only MIAs on overfitted DNNs.

\section{Discussion}\label{sec:Discussion}

\noindent 
\textbf{Stage-1 of Double-Dip beyond Label-only MIAs}: 
Our results in Sec. \ref{sec:evaluation} showed that Double-Dip was consistently effective in reducing ASR of an adversary carrying out label-only MIAs. 
We carry out experiments to examine the performance of Double-Dip against other types of MIAs. 
In particular, we look at entropy-based MIAs \cite{shokri2017membership}. 
Entropy-based MIAs distinguish members and nonmembers using magnitudes of a cross-entropy loss term. 
The cross-entropy loss under noise addition for a member sample will be smaller since they are typically farther away from a decision boundary, which results in a DNN model classifying them with higher confidence. 

Table \ref{tab:ResNet-18-Entropy} presents results when Stage-1 of Double-Dip is deployed against an adversary carrying out an entropy-based MIA \cite{shokri2017membership}. 
We examine the ASR and classification accuracy when (i) using Stage-1 of Double-Dip on a \emph{pretrained ResNet-18 model}, (ii) without transfer learning, and (iii) when using L1/ L2 regularization. 
We consider cases where an adversary carrying out an entropy-based MIA has (a) white-box, and (b) black-box model access. 
We examine CIFAR-10 and GTSRB target datasets of sizes $500$ and $1000$. 
For both datasets, transfer learning is effective in reducing ASR of an adversary carrying out a MIA while also achieving higher classification accuracy relative to no transfer learning or regularization. 

Stage-1 of Double-Dip is more effective against label-only MIAs than entropy-based MIAs (compare ASR values in Tables \ref{tab:ResNet-18-Target} and \ref{tab:ResNet-18-Entropy}). 
An explanation for this is an adversary carrying out an entropy-based MIA has precise information about confidence scores of model outputs (different than an adversary carrying out label-only MIAs who has access only to output labels). Such an adversary will thus be able to infer about membership more effectively. 
However, defenses against MIAs that use confidence scores of model outputs have been successfully developed, including in 
~\cite{jia2018attriguard, jia2019memguard, yang2020defending}.

\begin{table}[!h]
\centering
\caption{\textbf{Stage-1 of Double-Dip, Pretrained ResNet-18 Model, Entropy MIA}: Adversary success rate (ASR, \emph{lower is better}) and classification accuracy (ACC, \emph{higher is better}) for CIFAR-10 and GTSRB with training sets of sizes 500 and 1000 when an adversary carries out an entropy-based MIA. 
We compare (i) no transfer learning (NTL), (ii) regularization (L1, L2), and (iii) transfer learning (TL). TL-X indicates that X layers of the pretrained model are frozen. 
The best ASR and ACC values for a given training set size across both datasets is in \textbf{bold}; best ASR and ACC values in each cell are \underline{underlined}. 
Transfer learning yields lowest ASR while also ensuring significantly higher classification accuracy. 
(See \textbf{Sec. 7})
}\label{tab:ResNet-18-Entropy}
    \begin{tabular}{c|c|cc|cc}
        \toprule \hline
        \multicolumn{2}{c}{} & \multicolumn{2}{c}{\textbf{500}} & \multicolumn{2}{c}{\textbf{1000}}\\
        \hline
        \textbf{Dataset} & \textbf{Setting} & \%\textbf{ASR} & \%\textbf{ACC} & \%\textbf{ASR} & \%\textbf{ACC} \\
        \hline \hline
        \multirow{8}{*}{\thead{CIFAR-10}} & NTL & 92.7 &   36.2 &  93.5 & 43.0  \\
        & L1(0.001) & 91.8 & 36.4 & 86.0 & 41.5\\
        & L2(0.1) & 92.3   & 35.8 & 90.1 & 44.1 \\
        & \textbf{TL-0} & \cellcolor{gray!25} 82.9 & \cellcolor{gray!25} 65.8 & \cellcolor{gray!25} 79.5 &  \cellcolor{gray!25} 77.3 \\
        & \textbf{TL-20} & \cellcolor{gray!25} 79.0 &  \cellcolor{gray!25} 73.2 &   \cellcolor{gray!25}\underline{76.9}&\cellcolor{gray!25} 73.9 \\
        & \textbf{TL-40} & \cellcolor{gray!25} 78.8 & \cellcolor{gray!25} 76.2 &  \cellcolor{gray!25} 79.5 & \cellcolor{gray!25} 77.4 \\	
        & \textbf{TL-50} & \cellcolor{gray!25} \underline{78.3} &\cellcolor{gray!25} \underline{77.8} & \cellcolor{gray!25} 79.5 &  \cellcolor{gray!25} \underline{81.2}\\	
        & \textbf{TL-60} & \cellcolor{gray!25} 80.0 &  \cellcolor{gray!25} 72.6 &  \cellcolor{gray!25} 75.9 &\cellcolor{gray!25} 72.5 \\	
        \hline \hline 	

        \multirow{8}{*}{\thead{GTSRB}} & NTL & 78.6 &   42.0  &83.6& 61.4  \\
        & L1 (0.001) & 79.0 & 41.4 & 84.6& 58.9 \\
        & L2 (0.1) & 81.9  & 44.2& 86.0 & 60.3 \\
        & \textbf{TL-0} & \cellcolor{gray!25} 86.5& \cellcolor{gray!25} 60.6 & \cellcolor{gray!25}77.8&  \cellcolor{gray!25} 81.3\\
        & \textbf{TL-20} & \cellcolor{gray!25} 82.6&  \cellcolor{gray!25} 65.8  &   \cellcolor{gray!25}77.8&  \cellcolor{gray!25} 75.9 \\
        & \textbf{TL-40} & \cellcolor{gray!25} \underline{\textbf{69.2}}  & \cellcolor{gray!25} \underline{\textbf{89.2}}  & \cellcolor{gray!25} \underline{\textbf{66.3}} &   \cellcolor{gray!25} \underline{\textbf{92.3}}\\
        & \textbf{TL-50} & \cellcolor{gray!25} 83.4 & \cellcolor{gray!25} 78.2 & \cellcolor{gray!25}80.5  &  \cellcolor{gray!25} 82.9\\
        & \textbf{TL-60} & \cellcolor{gray!25} 86.7 & \cellcolor{gray!25} 64.2 & \cellcolor{gray!25}85.8&  \cellcolor{gray!25} 63.1\\

        \hline \bottomrule
     \end{tabular}
\end{table}

\noindent
\textbf{White-box vs. Black-box Model Access in Label-only MIAs}: 
Our results in Sec. \ref{sec:evaluation} indicate that an adversary carrying out a label-only MIA with black-box model access (using a SOTA query-efficient method, HopSkipJump \cite{chen2020hopskipjumpattack}) is able to achieve a higher ASR than an adversary with white-box model access (using a SOTA  computationally inexpensive gradient-based method, BIM \cite{BIM}) in a few cases. Our results are consistent with observations made in \cite{chen2020hopskipjumpattack}, where the authors showed that the HopSkipJump black-box adversarial learning technique outperformed multiple SOTA white-box adversarial learning baselines, including the BIM.

\begin{figure*}[!h]
    \centering
    \includegraphics[width=0.5\textwidth]{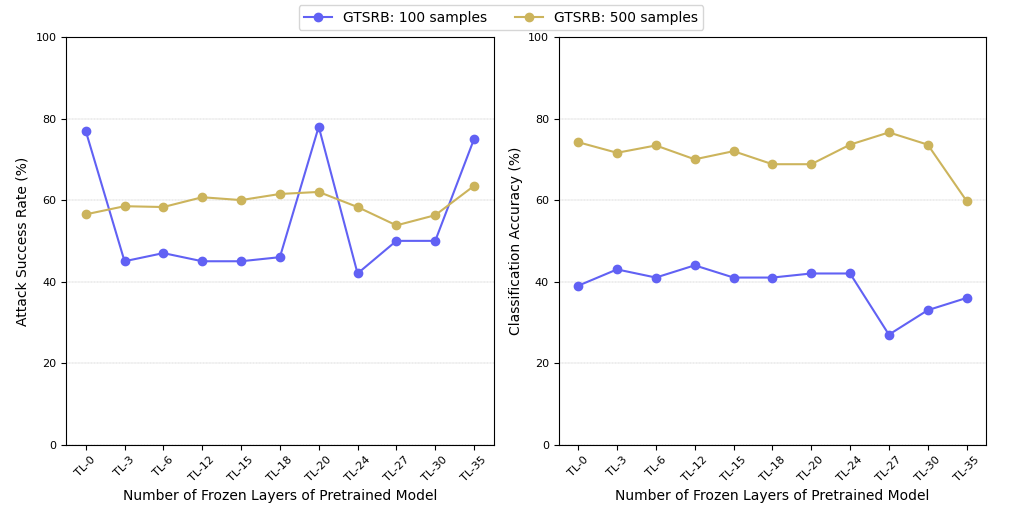}
    \caption{\textbf{Effect of Number of Frozen Layers}: This figure compares ASR (\emph{left}) and ACC (\emph{right}) values for different numbers of frozen layers of a pretrained VGG-19 model for the GTSRB dataset with training sets of sizes $100$ and $500$. A smaller training set size is more sensitive to the choice of the number of frozen layers of the pretrained model in Stage-1 of Double-Dip (compare fluctuations in ASR in left figure). 
    Fewer samples in the target dataset also enables achieving a lower ASR; however, this is accompanied by a reduction in classification accuracy. 
On the other hand, a larger-sized target dataset results in ASR and ACC values that are more or less independent of the choice of the number of frozen layers of the pretrained model. (See \textbf{Sec. 7}.)}
    \label{fig:vgg_difffrozenlayers_acc}
\end{figure*}

\noindent 
\textbf{Role of Number of Frozen Layers}: 
We investigate the effect of number of frozen layers of the pretrained model on ASR and ACC values. 
On one extreme, freezing few layers of the pretrained model will lead to transfer of a smaller number of weights and parameters, thereby limiting the ability to transfer features from pretrained to target model. 
On the other hand, freezing too many layers could result in scarcity of trainable layers leading to a lack of learnable weights that can adequately capture features of samples in the target dataset. 
We perform experiments on the GTSRB dataset to learn target models using a pretrained VGG-19 model in Fig. \ref{fig:vgg_difffrozenlayers_acc}. 

Our results indicate that the choice of the number of frozen layers is more critical when the size of the target dataset is smaller. 
This is underscored by larger fluctuations in ASR values when the target dataset has $100$ samples than when there are $500$ samples. 
Fewer samples in the target dataset also enables achieving a lower ASR in most cases. 
However, this is accompanied by a reduction in classification accuracy. 
On the other hand, a larger-sized target dataset results in ASR and ACC values that are more or less independent of the number of frozen layers of the pretrained model. 

\noindent
\textbf{No Correlated Features between Source and Target Datasets}: 
The choice of target and source datasets for our experiments in Sec. \ref{sec:evaluation} assumed that the presence of correlated features between the (TargetDataset, SourceDataset) pair. 
When the source and target datasets do not share features, the gap between source and target domains can be bridged by using techniques such as unsupervised domain adaptation methods \cite{long2016unsupervised}. 
One recently proposed mechanism that uses domain adaptation to obfuscate the training dataset, making it more challenging for an adversary to infer membership was used as a defense against MIAs in \cite{huang2021damia}. 
We believe that such domain adaptation techniques can be integrated with DoubleDip to defend against label-only MIAs when the target dataset size is small and when a pretrained model trained on a correlated dataset is not available. 
The integration of domain adaptation strategies with DoubleDip will enable a DNN model to generalize better across source and target domains, even in the absence of shared features. 
\section{Related Work
}\label{sec:RelatedWork}

\noindent{\bf Defenses against MIAs:} 
MIAs aim to determine whether a sample belongs to the training set of a classifier or not.  
There are three different approaches that an adversary with  black-box access to outputs of its target model can use to distinguish members and non-members. 
The adversary can 
(i) directly examine model outputs by studying confidence values or calculating loss values returned by the model~\cite{yeom2018privacy};  
(ii) learn a model to analyze differences between model outputs for members and nonmembers~\cite{shokri2017membership,ye2021shokri2}; or  
(iii) examine model outputs for perturbed variants of each input sample~\cite{PETS2021}. 
MIAs of this type presume that models behave differently for perturbed variants of members and nonmembers. 
However, all the above types of MIAs are confidence-score based. 
A new class of label-based or label-only MIAs was proposed in~\cite{choquette2021label, ccs2021} where an adversary only required knowledge of labels of samples rather than associated confidence scores. 

Two defenses, Memguard and Attriguard~\cite{jia2018attriguard, jia2019memguard, yang2020defending} were designed to protect against MIAs that use confidence scores of model outputs. 
These methods used the insight that perturbing the output of target models can mitigate impacts of confidence score-based MIAs. 
A defense against label-only MIAs was developed in \cite{rajabi2022ldl}, where an additive noise was used to create ambiguity to prevent a querying adversary from correctly determining whether a sample was a member or not. 
However, none of the above defenses assumed availability of a pretrained model that is trained on a (different) dataset from the dataset of interest to the user. \\

\noindent{\bf Transfer learning:} 
Transfer learning can be achieved through intra- (large model used to learn a smaller model for same dataset) or inter-domain (model trained on one dataset used to learn another model for different dataset) information transfer~\cite{zhuang2020comprehensive}. 
Depending on whether labels of samples from the target dataset are available or not, transfer learning has been successfully implemented on supervised~\cite{zhuang2015supervised} and unsupervised~\cite{bengio2012deep, cheplygina2019not, peng2016unsupervised} machine learning tasks. 
The authors of \cite{tu2020empirical, hendrycks2020pretrained} showed that using a pretrained model significantly improves robustness of the target model when the target model is trained on a different dataset. 
Transfer learning paradigms that use pretrained models~\cite{ribani2019Mainsurvey, simonyan2015very} typically require a large training set to be effective. 
Data augmentation has been proposed as a solution to increase the size of the training dataset~\cite{han2018new}. 
However, this solution was shown to increase the success of MIAs in~\cite{choquette2021label}. 
The authors of \cite{zou2020privacy} study MIAs in transfer learning when the pretrained model is assumed to be overfitted. 

In comparison, we consider publicly available pretrained models~\cite{he2016deep, simonyan2015very} that are not overfitted. 
We further assume that a user who wants to use the pretrained model to train a target model has access to a limited number of training samples. 
\section{Conclusion}\label{sec:conclussion}

We proposed \emph{Double-Dip}, a systematic empirical study of the role of transfer learning (TL) in thwarting label-only membership inference attacks (MIAs) on overfitted deep neural networks (DNNs). 
In Stage-1 of Double-Dip, we used TL to embed an overfitted DNN into a target model with shared feature space and parameter values similar to a pretrained model to reduce ASR of an adversary carrying out label-only MIAs while increasing classification accuracy. 
Stage-2 of Double-Dip employed randomization based on noise perturbation around an input to construct a region of constant output label to further reduce ASR, while maintaining classification accuracy. 
We used three (Target, Source) datasets ((CIFAR-10, ImageNet), (GTSRB, ImageNet), (CelebA, VGGFace2)) and four publicly available pretrained DNN models (VGG-19, ResNet-18, Swin-T, and FeceNet) to evaluate Double-Dip. 
Our experiments showed that Stage-1 was effective in reducing ASR 
and also significantly improved classification accuracy for nonmembers (up to $\mathbf{2.6}$\textbf{x} higher than SOTA regularization- and up to $\mathbf{3.4}$\textbf{x} higher than distillation-based methods). 
After Stage-2, ASR was reduced closer to $\mathbf{50\%}$, bringing it near to a random guess by the adversary. 
Our experiments showed efficacy of Double-Dip in thwarting label-only MIAs. 

\begin{acks}
This material is based upon work supported by the National Science Foundation under grant IIS 2229876 and is supported in part by funds provided by the National Science Foundation (NSF), by the Department of Homeland Security, and by IBM. Any opinions, findings, and conclusions or recommendations expressed in this material are those of the author(s) and do not necessarily reflect the views of the National Science Foundation or its federal agency and industry partners. 
This work is also supported by the Air Force Office of Scientific Research (AFOSR) through grant FA9550-23-1-0208, the Office of Naval Research (ONR) through grant N00014-23-1-2386, and the NSF through grant CNS 2153136.
\end{acks}
\bibliographystyle{ACM-Reference-Format}
\bibliography{example_paper}
\end{document}